\documentclass[amsmath,amssymb]{revtex4}
\usepackage{graphicx}
\usepackage{bm}

\newcommand{\be}{\begin{eqnarray}}
\newcommand{\ee}{\end{eqnarray}}

\newcommand{\ba}{\begin{array}}
\newcommand{\ea}{\end{array}}

\begin{document}
\title{
Geometric Analogue of Holographic Reduced Representation}
\author{Diederik Aerts $^1$, Marek Czachor $^{2,3}$, and Bart De Moor $^3$}
\affiliation{
$^1$ Centrum Leo Apostel (CLEA) and Foundations of the Exact Sciences (FUND)\\
Brussels Free University, 1050 Brussels, Belgium\\
$^2$ Katedra Fizyki Teoretycznej i Informatyki Kwantowej\\
Politechnika Gda\'nska, 80-952 Gda\'nsk, Poland\\
$^3$ ESAT-SCD, Katholieke Universiteit Leuven, 3001 Leuven, Belgium}

\begin{abstract}
Holographic reduced representations (HRR) are based on superpositions of convolution-bound $n$-tuples, but the $n$-tuples cannot be regarded as vectors since the formalism is basis dependent. This is why HRR cannot be associated with geometric structures. Replacing convolutions by geometric products one arrives at reduced representations analogous to HRR but interpretable in terms of geometry. Variable bindings occurring in both HRR and its geometric analogue mathematically correspond to two different representations of $Z_2\times\dots\times Z_2$ (the additive group of binary $n$-tuples with addition modulo 2). As opposed to standard HRR, variable binding performed by means of geometric product allows for computing exact inverses of all nonzero vectors, a procedure even simpler than approximate inverses employed in HRR.
The formal structure of the new reduced representation is analogous to cartoon computation, a geometric analogue of quantum computation.
\end{abstract}
\maketitle

\section{Introduction}

Reduced representations of cognitive structures are based essentially on two operations (binding and superposing) whose algebraic realizations vary from model to model. In \cite{Hinton}, where matrices representing roles act on vectors representing fillers, binding corresponds to matrix multiplication and superposition to vector addition. In tensor representations \cite{Smolensky} roles and fillers are represented by vectors which are bound by means of tensor products. The resulting simple tensors are superposed by addition. In holography-inspired memory models \cite{Gabor,Willshaw,Borsellino,Liepa,Murdock,Metcalfe,MetcalfeEich,Slack} binding is represented by convolution. Replacing tensor products by circular convolutions one arrives at holographic reduced representations (HRR) \cite{Plate95,Plate2003}. Restricting frequency-domain HRR to a subspace and switching to appropriately defined `logarithmic' variables one obtains binary spatter codes (BSC) \cite{Kanerva96,Kanerva97,Kanerva98}, with binary strings of length $n$ bound by $n$-dimensional sums mod~2 and superpositions modeled by majority-rule sums. Finally, in quantum computation (QC) \cite{Chuang} bits are bound into $n$-bit numbers by means of tensor products of two-dimensional complex vectors called qubits. QC is mathematically similar to tensor-product reduced representations, but differences occur at interpretational levels \cite{AC}.

Convolutions may be regarded as basis-dependent lossy compressions of the tensor product. The degree of compression can be estimated by means of dimensional analysis. In particular, circular convolutions occurring in HRR map pairs of $n$-tuples into $n$-tuples. In contrast, the ordinary convolution of two $n$-tuples is a $(2n-1)$-tuple, and an analogous tensor product would produce a $n^2$-tuple. These facts explain efficiency and usefulness of HRR in applications \cite{Plate2003}.

In spite of what one can often read in the literature, a convolution of two {\it vectors\/} is not well defined. This means that having two vectors, that is --- geometric objects, we cannot unambiguously identify a geometric object corresponding to their circular convolution. This seems to be a drawback, at least at the conceptual level. Geometry of some sort is at the roots of visualization, and visualization seems important for mathematical {\it understanding\/} and {\it proving\/} \cite{Widdows,WiddowsHiggins,Penrose,PG}. Hence the question: Is it possible to replace circular convolution by something similar but geometrically meaningful? If so, is there a relation to HRR?

We will argue that the most natural choice is to replace tensor products by {\it geometric products\/} \cite{Dorst2}, and not by circular convolutions. Geometric products, similarly to circular convolutions, preserve dimensionality at the level of {\it multivectors\/}. Multivectors are superpositions of {\it blades\/}, geometric-product analogues od simple tensors. Geometric products are also `exponents', in a sense that will be made precise later,  of $n$-dimensional sums mod~2. In consequence, geometric-product binding is in a unique relation to the binding employed in BSC, and the latter is a form of HRR.

Coding based directly on geometric products was recently applied to QC \cite{AC07,C07,AC07b} (a somewhat less direct way of linking geometric products with QC was used earlier in \cite{Somaroo}). As it turned out, all quantum algorithms of the standard formalism have geometric analogues. However, as opposed to standard QC that requires quantum mechanical implementations, the formalism based on geometric algebra (GA) requires {\it geometry\/} and not quantum mechanics. In principle, any system involving some geometry (Euclidean or not) is a candidate for implementation of a quantum algorithm. Systems where HRR are applicable might therefore, at least in principle, perform quantum algorithms.

The concept of GA is not new --- it appeared in the 19th century works of Grassmann \cite{G77} and Clifford \cite{Clifford} --- but geometric insights behind GA were forgotten for almost a century. At the end of 1960s the subject was revived with the works of Hestenes \cite{H1}. Today the Hestenes system \cite{HS,H2,H3} has found applications
(cf. \cite{Baylis,S,DDL,Pavsic,Doran,Lasenby96,robot,Bayro})
to topics as diverse as black holes, cosmology, quantum mechanics, quantum field theory, supersymmetry, beam dynamics, computer vision, robotics, protein folding, neural networks, computer aided design, and recently --- quantum computation. The link between HRR and GA suggests that the next step is to reformulate in a GA way the cognitive science.

The paper is organized as follows. Section~II introduces GA and its basic constructions (multivectors, invertibility, coding based on blades) and explains why GA naturally formalizes relations between geometric objects. In Section III we compare geometric product with circular convolution and explain why the latter cannot be interpreted in geometric terms. In Section IV we discuss the Fourier-space convolution algebra and show that it is in a one-to-one relation with the algebra of commuting matrices. In Section V we explain why variable binding in Fourier-space HRR is a representation, in group-theoretic sense, of the group $Z_2\times\dots\times Z_2$. Then, in Section VI we show that blades form a projective representation the same group and thus convolution binding is naturally represented in GA. We illustrate the new reduced representation in Section VII by reformulating the example Kanerva gave as an illustration of his BSC. Finally, in Section IX we reformulate the same example by means of a matrix representation of GA.

\section{Algebraic representation of geometric relations}

Consider the following set of relations $\approx$ involving two-dimensional basic shapes:
\be
--&\approx& |\,| \approx \Box\,\Box\approx1\\
-\,| &\approx& |\,-\approx \Box\\
-\,\Box &\approx& \Box\, - \approx |\\
|\,\Box &\approx& \Box\,| \approx -
\ee
One can think of them in at least two categories. One is simply a category of {\it understanding\/} relations between one- and two-dimensional objects: Square is formed from two orthogonal segments, a segment and a square imply which segment is missing, 1 means identity...  Another way of looking at these relations, more in the spirit of Grassmann and Clifford,  would be in terms of an {\it algebra\/} of geometric objects if associativity is assumed and 1 is treated as a neutral element.
Associativity then implies, for example,
$
-\,|\,-\approx \Box\,-\approx -\,\Box\approx|
$.
Let us make here a side remark that the `algebra' formalizes a procedure that resembles an IQ test.

The next step is to ask for higher-dimensional generalization and inclusion of {\it orientation\/}: Oriented line segments are vectors, plane segments have `sides', the relations between them will include a positive or negative sign, and we can also add cubes (having `insides' and `outsides'), their walls, and even higher dimensional structures.

Orientation makes the  algebra noncommutative
\be
\rightarrow\,\uparrow &\approx& +\Box\\
\uparrow\,\rightarrow &\approx& -\Box\\
\rightarrow\,\rightarrow &\approx& \uparrow\,\uparrow\,\approx 1
\ee
if we assume that `first up then right' generates the righthanded orientation, opposite to the lefthanded `first right then up'.
Adding both relations we find $\rightarrow\,\uparrow +\uparrow\,\rightarrow \approx 0$.
Taking three arrows and using associativity we obtain another, similar rule: $\Box\rightarrow+ \rightarrow\Box\approx 0$. The proof goes as follows
\be
\Box\rightarrow\,\approx\, \rightarrow\uparrow\rightarrow\, \approx\, \rightarrow(-\Box)\, \approx -\rightarrow\Box\,\approx\, \leftarrow\Box
\ee
However, we cannot simultaneously assume $\rightarrow\,\rightarrow \approx \uparrow\,\uparrow\,\approx 1$ and $\Box\Box\approx 1$. Indeed, anticommutativity of  $\rightarrow\,\uparrow \approx -\uparrow\,\rightarrow$ implies
\be
\Box\Box\approx \rightarrow\uparrow\rightarrow\uparrow\approx - \rightarrow\rightarrow\uparrow\uparrow\approx-11\approx -1
\ee
so we have to decide at which level to define the algebra. One can also think of this example as a hierarchy of geometric structures with increasing dimensions and different metric signatures. The first level is the pair $\{\rightarrow,\uparrow\}$ and the Euclidean-space signature is $(+,+)$. The second level is
$\{\bm 1,\rightarrow,\uparrow,\Box\}$ and the signature is $(+,+,+,-)$, known from space-time pseudo-Euclidean geometry.

The type of construction we have just outlined led Grassmann and Clifford to the algebra based on the concise formula
\be
b_k b_l+b_lb_k=2\delta_{kl} \bm 1\label{C}.
\ee
Here the $b$s denote orthonormal basis vectors in some $n$-dimensional real Euclidean space, $\bm 1$ is the neutral element of the algebra, and $\delta_{kl}$ is the Kronecker delta. The 2D plane example is reconstructed from (\ref{C}) if
$b_1=\rightarrow$, $b_2=\uparrow$, $b_1b_2=\Box=b_{12}$.

The algebra (\ref{C}) is known as the Clifford algebra and may be regarded as the grammar of GA. It refers to a concrete basis, but can be reformulated in a basis-free way. Indeed, consider two vectors $x=\sum_{k=1}^nx_kb_k$ and $y=\sum_{k=1}^ny_kb_k$. Their geometric product reads
\be
xy
&=&
\underbrace{\sum_{k=1}^nx_ky_k \bm 1}_{x\cdot y} +\underbrace{\sum_{k<l}(x_ky_l-y_kx_l)b_kb_l}_{x\wedge y}
\ee
The geometric product $xy$ is a sum of two terms. The {\it scalar\/} $x\cdot y=y\cdot x$ is known as the {\it inner product\/}. The {\it bivector\/}
$x\wedge y=-y\wedge x$ is the {\it outer product\/}. In 3D the length of $x\wedge y$ represents the area of the parallelogram spanned by $x$ and $y$.

In arbitrary dimension the bivector $x\wedge y$ represents an oriented plane segment. Grassmann and Clifford introduced geometric product by means of the basis-independent formula involving a {\it multivector\/}
\be
xy=x\cdot y+x\wedge y \label{AB}
\ee
which implies (\ref{C}) when restricted to the orthonormal basis. Inner and outer product can be defined directly from $xy$:
\be
x\cdot y &=& \frac{1}{2}(xy+yx),\\
x\wedge  y &=& \frac{1}{2}(xy-yx).
\ee
The most ingenious element of (\ref{AB}) is that it adds two apparently different objects: A scalar and a plane element. This seems `wrong' but this is precisely what happens when we speak of complex numbers or extend space and time to space-time. Apparently, the person to be blamed for the fact that multivectors may nowadays seem weird is Gibbs \cite{Gibbs}, who was more famous at the time than Grassmann or Clifford, and spoiled their work by separating the geometric product into two separate operations --- losing associativity and invertibility, as we shall see shortly.

Geometric interpretation and visualization of multivectors can be formulated in various ways, and various interpretations can be found in the literature. Perhaps the easiest way of getting used to thinking in GA terms is to browse through the websites devoted to GA (cf. \cite{lomont.org}).
The approach we found useful for multi-bit problems of QC was a representation in terms of directed colored polylines \cite{AC07b}.

Geometric product for vectors $x$, $y$, $z$ can be defined by the following rules:
\be
(xy)z &=& x(yz),\\
x(y+z) &=& xy+xz,\\
(x+y)z &=& xz+yz,\\
x^2 &=& |x|^2,
\ee
where $|x|$ is a positive scalar called the magnitude of $x$. The rules imply that $x\cdot y$ must be a scalar since
\be
xy+yx=|x+y|^2-|x|^2-|y|^2.
\ee
GA allows to speak of inverses of vectors: $x^{-1}=x/|x|^2$. $x$ is invertible (i.e. possesses an inverse) if its magnitude is nonzero. Geometric product of an arbitrary number of invertible vectors is also invertible. The possibility of inverting all nonzero-magnitude vectors is perhaps the most important difference between GA and tensor or convolution algebras.

Geometric products of {\it different\/} basis vectors
\be
b_{k_1\dots k_j}=b_{k_1}\dots b_{k_j},
\ee
$k_1<\dots <k_j$, are called blades. In $n$-dimensional (pseudo-)Euclidean space there are $2^n$ different blades. This can be seen as follows. Let $\{x_1,\dots, x_n\}$ be a sequence of bits. Blades in an $n$-dimensional space can be written as
\be
c_{x_1\dots x_n}=b_{1}^{x_1}\dots b_{n}^{x_n}
\ee
where $b_k^0=\bm 1$, which shows that blades are in a one-to-one relation with $n$-bit numbers. This observation is at the roots of the GA reformulation of QC introduced in \cite{AC07}. A general multivector is a linear combination of blades,
\be
\psi
&=&
\sum_{x_1\dots x_n=0}^1\psi_{x_1\dots x_n}c_{x_1\dots x_n}, \label{psi}
\ee
with real coefficients $\psi_{x_1\dots x_n}$ \footnote{A complex structure, if needed, may be introduced by means of an additional bit without any need of complex numbers. This is how the `imaginary unit' $i$ was used in \cite{C07} in the context of quantum gates. It must be stressed, however, that in the GA literature it is a tradition to replace $i$ by a blade. For example, we have seen that in 2D $(b_{1}b_2)^2=-\bm 1$ which explains why $b_1b_2$ has properties analogous to $i$. Nevertheless, this construction does not properly work in our context (cf. the discussion in \cite{AC07b}) and we have to use a different representation of $i$ --- equivalent to a $\pi/2$ rotation in 2D. These subtleties are not important for the discussion given in the present paper.}.

An inverse of a multivector is a well defined notion but not all multivectors are invertible. To find an inverse of a multivector is not an entirely trivial task in general. It resembles an analogous problem of inverting matrices. But all blades and geometric products of invertible vectors are invertible.

\section{Circular convolution vs. geometric product}

We have mentioned in Section~I that
convolutions are not defined on vectors but only on $n$-tuples. Let us explain this statement in more detail on the example of circular convolution.
Circular convolution $x\circledast y$ of the $n$-tuples $x=(x_0,\dots ,x_{n-1})$, $y=(y_0,\dots ,y_{n-1})$ is defined as
\be
(x\circledast y)_j
&=&
\sum_{k=0}^{n-1}x_ky_{j-k\, {\rm mod}\,n}.\label{conv}
\ee
For pairs the formula (\ref{conv}) reads explicitly
\be
\left(
\begin{array}{c}
x_0\\
x_1
\end{array}
\right)
\circledast
\left(
\begin{array}{c}
y_0\\
y_1
\end{array}
\right)
&=&
\left(
\begin{array}{c}
x_0y_{0}+x_1y_{1}
\\
x_0y_{1}+x_1y_{0}
\end{array}
\right).\label{conv1}
\ee
Let us note that if we tried to interpret (\ref{conv1}) in terms of {\it vectors\/} we would have to implicitly assume that the pairs on both sides of (\ref{conv1}) correspond to the {\it same\/} basis (otherwise the formula would be completely ambiguous). So let us take two different bases
$\{\bm b_0,\bm b_1\}$ and $\{\bm b'_0,\bm b'_1\}$, and two vectors $\bm x$, $\bm y$. Each of these vectors can be written in both bases:
\be
\bm x &=& x_0 \bm b_0+x_1 \bm b_1=x'_0 \bm b'_0+x'_1 \bm b'_1,\\
\bm y &=& y_0 \bm b_0+y_1 \bm b_1=y'_0 \bm b'_0+y'_1 \bm b'_1.
\ee
Circular convolutions of vectors would be meaningful if in any two bases we would find
\be
\bm x\circledast \bm y
&=&
(x_0y_{0}+x_1y_{1})\bm b_0
+
(x_0y_{1}+x_1y_{0})\bm b_1
=
(x'_0y'_{0}+x'_1y'_{1})\bm b'_0
+
(x'_0y'_{1}+x'_1y'_{0})\bm b'_1
\ee
which is not the case. Indeed, let us take
the basis rotated by $\pi/2$ ($\bm b'_0=\bm b_1$, $\bm b'_1=-\bm b_0$, $x'_0=x_1$, $x'_1=-x_0$, $y'_0=y_1$, $y'_1=-y_0$)
\be
\bm x &=& x'_0\bm b'_0+x'_1\bm b'_1=x_1\bm b_1+(-x_0)(-\bm b_0),\\
\bm y &=& y'_0\bm b'_0+y'_1\bm b'_1=y_1\bm b_1+(-y_0)(-\bm b_0),
\ee
implying
\be
(x'_0y'_{0}+x'_1y'_{1})\bm b'_0
+
(x'_0y'_{1}+x'_1y'_{0})\bm b'_1
&=&
(x_1y_{0}+x_0y_{1})\bm b_0
+
(x_0y_{0}+x_1y_{1})\bm b_1\\
&\neq &
(x_0y_{0}+x_1y_{1})\bm b_0
+
(x_0y_{1}+x_1y_{0})\bm b_1.
\ee
In contrast, for the geometric product we find
\be
\bm x\bm y
&=&
(x_0y_{0}+x_1y_{1})\bm 1
+
(x_0y_{1}-x_1y_{0})\bm b_0\bm b_1
=
(x'_0y'_{0}+x'_1y'_{1})\bm 1
+
(x'_0y'_{1}-x'_1y'_{0})\bm b'_0\bm b'_1,
\ee
which does not depend on the choice of basis. Let us note that the difference between geometric product and circular convolution boils down in this example to a single change of sign and reshuffling of components. To understand the latter property we have to bear in mind that the GA of a 2D plane is $2^2$-dimensional. A general multivector is here of the form $\psi=\alpha\bm 1+\beta \bm b_0+\gamma \bm b_1+\delta \bm b_0\bm b_1$. Our calculation, in terms of the 4-tuples $(\alpha,\beta,\gamma,\delta)$, reads
\be
\left(
\begin{array}{c}
0\\
x_0\\
x_1\\
0
\end{array}
\right)
\left(
\begin{array}{c}
0\\
y_0\\
y_1\\
0
\end{array}
\right)
=
\left(
\begin{array}{c}
x_0y_{0}+x_1y_{1}\\
0\\
0\\
x_0y_{1}-x_1y_{0}
\end{array}
\right).
\ee
There do exist matrix representations of GA (cf. Section IX).
It is known, however, that matrix representations of GA are not the most efficient implementations of GA-based algorithms. The algorithms that work efficiently in practice are based on direct calculations performed in terms of the GA rules (cf. \cite{gable}).

\section{Circular convolution in the Fourier space vs. matrix algebra}

A general multivector (\ref{psi}) can be represented by the $2^n$-tuple $(\psi_{0_1\dots 0_n},\dots,\psi_{1_1\dots 1_n})$.
The neutral element $\bm 1$ corresponds in this notation to $(1,0,\dots,0)$.

Similarly, the neutral element of the $\circledast$-algebra of $n$-tuples is the $n$-tuple $I=(1,0,\dots,0)$.
By definition, the $\circledast$-inverse $x^{-1}$ of the $n$-tuple $x$ satisfies $x^{-1}\circledast x=I$. An important map is the `involution'
\be
(x^*)_j = x_{-j\, {\rm mod}\,n}.
\ee
Let us now rewrite $\circledast$ and $\ast$ in the Fourier space. The Fourier transform of $(x_0,\dots,x_{n-1})$,
\be
\hat x_k &=& \sum_{l=0}^{n-1}x_l e^{-2\pi i kl/n},
\ee
satisfies
\be
(\widehat{x\circledast y})_k &=& \hat x_k\hat y_k,\\
(\widehat{x^*})_k &=& \overline{\hat x_k},\\
(\widehat{x^*\circledast y})_k &=& \overline{\hat x_k}\hat y_k,
\ee
where $\overline{\hat x_k}$ denotes complex conjugation.
The Fourier transform of $I=(1,0,\dots,0)$ is $\hat I=(1,1,\dots,1)$. Thus
\be
(\widehat{x^{-1}})_k &=& \frac{1}{\hat x_k}.
\ee
$x$ is not $\circledast$-invertible if any component of its Fourier transform is 0. This is why exact inverses are not used in standard HRR. Plate explains in \cite{Plate2003} why  $x^*$ may be regarded as an approximate inverse of $x$, and why in the presence of noise --- a generic situation in HRR --- application of exact inverses would lead to unstable algorithms. Only in the case of unitary $x$ (i.e. such that $x^*=x^{-1}$,  $|\hat x_k|=1$) the approximate inverse is exact. In GA, in contrast to HRR, exact inverses of nonzero vectors always exist, do not lead to instabilities, and are easy to calculate since $x^{-1}=x/|x|^2$.

Circular convolution in the Fourier space is thus equivalent to multiplication of diagonal matrices. Accordingly, the $\circledast$-inverse is the matrix inverse, and involution means Hermitian conjugation. The notion of $\circledast$-unitarity coincides with matrix unitarity. Fourier-space HRR involves binding represented by the matrix product of diagonal matrices and superposition is performed by matrix addition.
HRR is implicitly an operator procedure but involving only commuting operators. These operators are in general neither Hermitian nor unitary but have the property of being {\it normal\/}, i.e. commute with their Hermitian conjugates.

Plate mentions in \cite{Plate2003} (Section 3.6.7) that commutativity can cause ambiguities, so
certain noncommutative variants of $\circledast$ may be in principle considered. For example, combination of $\circledast$ with permutations of components introduces noncommutativity for the price of associativity. Still another alternative mentioned in \cite{Plate2003} is to work with vectors that can be written as matrices (i.e. with $n=m^2$, for some $m$) and use matrix multiplication. Apparently, this kind of reduced representation has not been studied in the literature so far.

Our claim is that the GA formalism is a natural noncommutative alternative to HRR, but to appreciate it we have to go deeper into the structure of the Fourier-space HRR.

\section{From Fourier-space HRR to BSC}

Consider a general HRR-type Fourier-space superposition of $N$ diagonal matrices $U_j$:
$\psi = \sum_{j=1}^N  U_j$.
Now let us restrict $U_j$ to matrices of the form $U_j=e^{i\pi P_{x_j}}=(-1)^{P_{x_j}}$ where $P_{x_j}$ are diagonal matrices whose only nonzero elements are equal to 1. In other words, a diagonal of $P_{x_j}$ is a sequence of bits: $P_{x_j}={\rm diag}(x_{j,1},\dots,x_{j,n})$, $x_{j,k}=0,1$. Such a $P_{x_j}$ is a projector: $P_{x_j}^2=P_{x_j}$. Under these restrictions the diagonal unitary matrix $U_j$ has the diagonal consisting of $(-1)^{x_{j,k}}=\pm 1$. The next step is to consider the new diagonal matrix
$
\Psi = {\rm sign}(\psi)
$
defined via the spectral theorem from
\be
{\rm sign}(x)
&=&
\left\{
\begin{array}{ll}
+1 & {\rm for\,}x\geq 0\\
-1 & {\rm for\,}x< 0\\
\end{array}
\right.
.
\ee
$\Psi$ is again a unitary diagonal matrix whose only nonzero elements are equal to $\pm 1$ and hence can be written as
$\Psi=e^{i\pi P_x}=(-1)^{P_x}$.
$P_x={\rm diag}(x_{1},\dots,x_{n})$ is a new projector, i.e. a diagonal matrix with bits on the diagonal.
In effect, we have produced
a new binary sequence $(x_{1},\dots,x_{n})$ from a collection of $N$ binary sequences $(x_{j,1},\dots,x_{j,n})$. The relevant formula (majority-rule summation) reads
\be
x_k &=& \boxplus_{j=1}^N x_{j,k}=\Theta\Big(\frac{1}{N}\sum_{j=1}^N x_{j,k}-\frac{1}{2}\Big)
\ee
where
\be
{\Theta}(x)
&=&
\left\{
\begin{array}{ll}
1 & {\rm for\,}x\geq 0\\
0 & {\rm for\,}x< 0\\
\end{array}
\right.
\ee
is the Heaviside step function.

The final step leading to BSC is to treat each $U_j$ as a product of two unitary diagonal matrices $R_j$ and $F_j$, also consisting of pluses and minuses on diagonals,
\be
\Psi &=& {\rm sign}(\psi)=(-1)^{P_x}= {\rm sign}\big(\sum_{j=1}^N  R_jF_j\big)
={\rm sign}\big(\sum_{j=1}^N  (-1)^{P_{x_j}}(-1)^{P_{y_j}}\big)
=
{\rm sign}\big(\sum_{j=1}^N  (-1)^{(P_{x_j}\oplus P_{y_j})}\big)
.
\ee
$R_j$, $P_{x_j}$ represent roles, $F_j$, $P_{y_j}$ are fillers, and $P_{x_j}\oplus P_{y_j}$ denotes matrix addition mod~2.

Since $P_{x_j}$ and $P_{y_j}$ are diagonal matrices with 0s and 1s on the diagonals, say, $P_{x_j}={\rm diag}(x_{j,1},\dots,x_{j,n})$, $P_{y_j}
={\rm diag}(y_{j,1},\dots,y_{j,n})$, the map
\be
(x_{1},\dots,x_{n})
&=&
\boxplus_{j=1}^N (x_{j,1},\dots,x_{j,n})\oplus (y_{j,1},\dots,y_{j,n})
\label{bsc}
\ee
is the thresholded majority-rule componentwise addition of $n$-dimensional sums mod 2 ($n$-dimensional exclusive alternatives, XORs).

Now consider a single bit $X$ and a sequence of $N$ bits $(x_1, \dots, x_N)$.
Then the following distributivity of $\oplus$ over $\boxplus$ holds true:
\be
X\oplus \big(\boxplus_{j=1}^N x_j\big) &=& \boxplus_{j=1}^N \big(X\oplus x_j\big).\label{distr}
\ee
(\ref{distr}) naturally generalizes to the $n$-dimensional variants of $\boxplus$ and XOR.
Eq.~(\ref{bsc}) is the BSC where binary strings are bound by $\oplus$ and superposed by thresholded addition. Decoding of information is based in BSC on (\ref{distr}).

Let us rephrase the main result as follows. Circular convolution of $n$-tuples whose entries consist of $\pm 1$ may be regarded as a multiplicative representation of XOR of $n$-bit strings, and the appropriate map is
\be
x\oplus y\mapsto (-1)^{(P_x\oplus P_y)}=(-1)^{P_x}(-1)^{P_y}=RF.
\ee
The link between HRR and BSC is given by the map $x\mapsto (-1)^{P_x}$, where $x$ on its left-hand side is a binary string of length $n$ while on the right-hand side $x$ occurs at the diagonal of an $n\times n$ matrix $P_x$.

In the next Section we will see that geometric product represents the {\it same\/} structure but in a {\it projective\/} way.

\section{Geometric product as a projective representation of XOR}

Let $x_1\dots x_n$ and $y_1\dots y_n$ be binary representations of two $n$-bit numbers $x$ and $y$.
Now let us consider two blades $c_x=c_{x_1\dots x_n}=b_1^{x_1}\dots b_n^{x_n}$, $c_y=c_{y_1 \dots y_n}=b_1^{y_1}\dots b_n^{y_n}$. We will show that geometric product of $c_x$ and $c_y$ equals, up to a sign, $c_{x\oplus y}$. In this sense the map $x\mapsto c_x$ is an analogue of the exponential map $x\mapsto (-1)^{P_x}$.

Let us begin with examples:
\be
b_1b_1
&=&
c_{10\dots 0}c_{10\dots 0}
=
1=c_{0\dots 0}=c_{(10\dots 0)\oplus (10\dots 0)}\\
b_1b_{12}
&=&
c_{10\dots 0}c_{110\dots 0}
=
b_1b_1b_2=b_2=c_{010\dots 0}=c_{(10\dots 0)\oplus (110\dots 0)}\\
b_{12}b_1
&=&
c_{110\dots 0}c_{10\dots 0}
=
b_1b_2b_1=-b_2b_1b_1=-b_2=-c_{010\dots 0}=-c_{(110\dots 0)\oplus(10\dots 0)}\\
b_{1257}b_{26}
&=&
c_{11001010\dots 0}c_{0100010\dots 0}
=
b_1b_2b_5b_7b_2b_6\nonumber\\
&=&
(-1)^3b_1b_5b_6b_7
=
(-1)^3
c_{10001110\dots 0}
=
(-1)^D
c_{(11001010\dots 0)\oplus(0100010\dots 0)}.
\ee
The number $D$ is the number of times a 1 from the right string had to ``jump" over a 1 from the left one during the process of shifting the right string to the left.

The above observations, generalized to arbitrary strings of bits, yield
\be
c_{x_1\dots x_n}c_{y_1\dots y_n}
&=&
(-1)^{\sum_{k<l}y_kx_l}c_{(x_1\dots x_n)\oplus(y_1\dots y_n)}.
\label{GAr}
\ee
Indeed, for two arbitrary strings of bits we have
\be
D=y_1(x_2+\dots+x_n)+y_2(x_3+\dots+x_n)+\dots+y_{n-1}x_n=\sum_{k<l}y_kx_l.
\ee
We conclude that the map
\be
(x_1,\dots, x_n)\times(y_1,\dots, y_n)
\mapsto
(x_1,\dots, x_n)\oplus(y_1,\dots, y_n)=(x_1\oplus y_1,\dots, x_n\oplus y_n)
\ee
is projectively (i.e. up to a sign) represented in GA by means of (\ref{GAr}).

Representations `up to a sign' play an important role in physics and are at the roots of many phenomena such as half-integer spin and fermionic statistics. To the best of our knowledge the link between projective representations of XOR and GA was noticed for the first time in the preliminary version of this paper \cite{GABSC}

\section{Geometric analogue of HRR}

Let us begin with illustrating the original BSC by means of the example taken from \cite{Kanerva97}. The records are represented by unstructured randomly chosen strings of bits. The encoded record is
\be
{\bf PSmith}
&=&
({\bf name}
\oplus
{\bf Pat})
\boxplus
({\bf sex}
\oplus
{\bf male})
\boxplus
({\bf age}
\oplus
{\bf 66}).
\ee
Decoding of the ``name" looks as follows
\be
{\bf Pat'}
&=&
{\bf name}
\oplus
{\bf PSmith}\nonumber\\
&=&
{\bf name}
\oplus
\big[
({\bf name}
\oplus
{\bf Pat})
\boxplus
({\bf sex}
\oplus
{\bf male})
\boxplus
({\bf age}
\oplus
{\bf 66})
\big]
\nonumber\\
&=&
{\bf Pat}
\boxplus
({\bf name}
\oplus
{\bf sex}
\oplus
{\bf male})
\boxplus
({\bf name}
\oplus
{\bf age}
\oplus
{\bf 66})
\nonumber\\
&=&
{\bf Pat}
\boxplus
{\rm noise}
\to {\bf Pat}.
\ee
We have used here the involutive nature of XOR and the fact that the ``noise" can be eliminated by clean-up memory. The latter means that we compare ${\bf Pat'}$ with records stored in some memory and check, by means of the Hamming distance, which of the stored elements is closest to ${\bf Pat'}$. A similar trick could be done be means of circular convolution in HRRs, but then we would have used the  inverse {\bf name}$^{-1}$ or the involution {\bf name}$^*$, and an appropriate measure of distance. Again, the last step is comparison of the noisy object with the ``pure" objects stored in clean-up memory. This is how standard BSC works.

We can now use the exponential map $x\mapsto (-1)^{P_x}$ to turn BSC into HRR. Let us, however, proceed in the geometric way and employ $x\mapsto c_x$.
The roles and fillers are represented by randomly chosen blades:
\be
{\bf PSmith}
&=&
{\bf name}\cdot
{\bf Pat}
+
{\bf sex}\cdot
{\bf male}
+
{\bf age}\cdot
{\bf 66}.
\ee
The dot ``$\cdot$" is the geometric product and $\boxplus$ is replaced, similarly to HRR, by ordinary addition.
At the level of explicit blades the record corresponds to the multivector {\bf PSmith}
\be
{\bf PSmith}
&=&
c_{a_1\dots a_n}c_{x_1\dots x_n}
+
c_{b_1\dots b_n}c_{y_1\dots y_n}
+
c_{c_1\dots c_n}c_{z_1\dots z_n}.
\ee
The blades indexed by the beginning of the alphabet represent roles (name, sex, age) while the remaining ones correspond to the fillers (Pat, male, 66). The decoding looks as follows
\be
{\bf name}\cdot{\bf PSmith}
&=&
c_{a_1\dots a_n}
\big[
c_{a_1\dots a_n}c_{x_1\dots x_n}
+
c_{b_1\dots b_n}c_{y_1\dots y_n}
+
c_{c_1\dots c_n}c_{z_1\dots z_n}\big]\\
&=&
\pm c_{x}
\pm
c_{a\oplus b\oplus y}
\pm
c_{a\oplus c\oplus z}\\
&=&
\pm {\bf Pat}
+
{\rm noise}.
\ee
The signs have to computed by means of (\ref{GAr}).

An analogue of clean-up memory can be constructed in various ways. One possibility is to make sure that fillers, $c_{x}$ etc. are orthogonal to the noise term. For example, let us take the fillers of the form $c_{x_1\dots x_{k}0\dots 0}$, where the first $k\ll n$ bits are selected at random, but the remaining $n-k$ bits are all 0. Let the roles be taken, as in Kanerva's BSC, with {\it all\/} the bits generated at random.
The term $c_{(x_1\dots x_n)\oplus(y_1\dots y_n)\oplus(y_1\dots y_n)}$ will with high probability contain at least one $y_j=1$, $k<j\leq n$, and thus will be orthogonal to the fillers. The clean-up memory will consist of vectors with $y_j=0$, $k<j\leq n$, i.e of the filler form.

\section{Binary coding in different bases}

In quantum mechanics bits can be associated with any set of qubits (i.e. with any basis in a 2-dimensional complex space). The freedom to choose the basis is crucial for quantum cryptography, but is generally not used in QC. In QC one typically works in the so-called computational basis, which is fixed in advance. This, of course, does not change the fact that the whole formalism of QC is based on vectors and not on $n$-tuples.

A similar situation occurs in our GA analogue of HRR. A multivector $\psi$ (\ref{psi}) is a combination of blades, and blades represent binary numbers only when we fix the basis. To put it differently, the same single multivector $\psi$ can be associated with different reduced representations, but whether this freedom is of any practical use is an open question.

\section{Cartan representation of Clifford algebras}

It is useful to be able to work with matrix representations of GA. Although this is not an efficient way of doing GA computations, matrix representations allow to perform independent cross-checks of various GA constructions and algorithms.
In this section we give an explicit matrix representation of GA. We begin with Pauli's matrices
\be
\sigma_1
=
\left(
\begin{array}{cc}
0 & 1\\
1 & 0
\end{array}
\right),
\quad
\sigma_2
=
\left(
\begin{array}{cc}
0 & -i\\
i & 0
\end{array}
\right)
,
\quad
\sigma_3
=
\left(
\begin{array}{cc}
1 & 0\\
0 & -1
\end{array}
\right).
\ee
GA of a plane is represented as follows:  $1=2\times 2$ unit matrix, $b_1=\sigma_1$,
$b_2=\sigma_2$, $b_{12}=\sigma_1\sigma_2=i\sigma_3$. Alternatively, we can write
$c_{00}=1$, $c_{10}=\sigma_1$, $c_{01}=\sigma_2$,
$c_{11}=i\sigma_3$, and
\be
\psi_{00} c_{00}+\psi_{10} c_{10}+\psi_{01} c_{01}+\psi_{11} c_{11}
=
\left(
\begin{array}{cc}
\psi_{00} +i\psi_{11} & \psi_{10} -i\psi_{01}\\
\psi_{10} +i\psi_{01} & \psi_{00}-i\psi_{11}
\end{array}
\right).
\ee
This is equivalent to encoding $2^2=4$ real numbers into two complex numbers.

In 3-dimensional space we have
$1=2\times 2$ unit matrix, $b_1=\sigma_1$,
$b_2=\sigma_2$, $b_{3}=\sigma_3$, $b_{12}=\sigma_1\sigma_2=i\sigma_3$,
$b_{13}=\sigma_1\sigma_3=-i\sigma_2$,
$b_{23}=\sigma_2\sigma_3=i\sigma_1$,
$b_{123}=\sigma_1\sigma_2\sigma_3=i$.

Now the representation of
\be
\sum_{ABC=0,1}\psi_{ABC}c_{ABC}
=
\left(
\begin{array}{cc}
\psi_{000} +i\psi_{111} + \psi_{001}+i \psi_{110},
&
\psi_{100}+i\psi_{011} -i\psi_{010}-\psi_{101}\\
\psi_{100}+i\psi_{011} +i\psi_{010}+\psi_{101},
&
\psi_{000}+i\psi_{111}
-\psi_{001}-i \psi_{110}
\end{array}
\right)
\ee
is equivalent to encoding $2^3=8$ real numbers into 4 complex numbers.

An arbitrary $n$-bit record can be encoded into the matrix algebra known as Cartan's representation of Clifford algebras \cite{BT}:
\be
b_{2k}
&=&
\underbrace{\sigma_1\otimes\dots\otimes \sigma_1}_{n-k}
\otimes\,\sigma_2\otimes
\underbrace{1\otimes\dots\otimes 1}_{k-1},\\
b_{2k-1}
&=&
\underbrace{\sigma_1\otimes\dots\otimes \sigma_1}_{n-k}
\otimes\,\sigma_3\otimes
\underbrace{1\otimes\dots\otimes 1}_{k-1}.
\ee
So let us return to the example of Pat Smith. For simplicity take $n=4$ so that we can choose the representation
\be
\left.
\begin{array}{rcl}
{\bf Pat} &=& c_{1100},\\
{\bf male} &=& c_{1000},\\
{\bf 66} &=& c_{0100},
\end{array}
\right\}{\rm fillers}
\ee
\be
\left.
\begin{array}{rcl}
{\bf name} &=& c_{1010},\\
{\bf sex} &=& c_{0111},\\
{\bf age} &=& c_{1011}.
\end{array}
\right\}{\rm roles}
\ee
The fillers have only the first two bits selected at random, the last two are 00.
The roles are numbered by randomly selected strings of bits.

The explicit matrix representations are:
\be
{\bf Pat} &=& c_{1100}=b_1b_2
=
(\sigma_1\otimes\sigma_1\otimes \sigma_1\otimes\,\sigma_3)
(\sigma_1\otimes\sigma_1\otimes \sigma_1\otimes\,\sigma_2)
=
1\otimes 1\otimes 1\otimes (-i\sigma_1)
\\
{\bf male} &=& c_{1000}=b_1
=
\sigma_1\otimes\sigma_1\otimes \sigma_1
\otimes\,\sigma_3\\
{\bf 66} &=& c_{0100}
=
b_2
=
\sigma_1\otimes\sigma_1\otimes \sigma_1\otimes\,\sigma_2\\
{\bf name} &=& c_{1010}
=
b_1b_3
=
(\sigma_1\otimes\sigma_1\otimes \sigma_1\otimes\,\sigma_3)
(\sigma_1\otimes\sigma_1\otimes\,\sigma_3\otimes 1)\nonumber\\
&=&
1\otimes 1\otimes (-i\sigma_2)\otimes\,\sigma_3
\nonumber\\
{\bf sex} &=& c_{0111}
=
b_2b_3b_4
=
(\sigma_1\otimes\sigma_1\otimes \sigma_1\otimes\,\sigma_2)
(\sigma_1\otimes\sigma_1\otimes\,\sigma_3\otimes 1)
(\sigma_1\otimes\sigma_1\otimes\,\sigma_2\otimes 1)\nonumber\\
&=&
\sigma_1\otimes\sigma_1\otimes (-i 1)\otimes\,\sigma_2,\\
{\bf age} &=& c_{1011}=b_1b_3b_4
=
(\sigma_1\otimes\sigma_1\otimes \sigma_1\otimes\,\sigma_3)
(\sigma_1\otimes\sigma_1\otimes\,\sigma_3\otimes 1)
(\sigma_1\otimes\sigma_1\otimes\,\sigma_2\otimes 1)\nonumber\\
&=&
\sigma_1\otimes\sigma_1\otimes (-i 1)\otimes\,\sigma_3
\ee
The whole record --- where the superposition is taken for clarity with arbitrary parameters $\alpha$, $\beta$, and $\gamma$ --- reads
\be
{\bf PSmith}
&=&
\alpha\,{\bf name}\cdot{\bf Pat}
+
\beta\,{\bf sex}\cdot{\bf male}
+
\gamma\, {\bf age}\cdot
{\bf 66}\nonumber\\
&=&
\alpha c_{1010}c_{1100}
+
\beta c_{0111}c_{1000}
+
\gamma c_{1011}c_{0100}\nonumber\\
&=&
\alpha c_{0110}
-
\beta c_{1111}
+
\gamma c_{1111}\nonumber
\ee
The two noise terms are here linearly dependent by accident.
This is a consequence of too small dimensionality of our binary strings (four bits, whereas in realistic cases Kanerva suggested $10^4$ bit strings). This is the price we pay for simplicity of the example. Decoding the name involves two steps. First
\be
{\bf name}\cdot{\bf PSmith}
&=&
c_{1010}\cdot{\bf PSmith}
=
c_{1010}
\big[
\alpha c_{0110}
-
(\beta-\gamma) c_{1111}
\big]\nonumber\\
&=&
-
\alpha c_{1100}
-
(\beta-\gamma) c_{0101}
\nonumber\\
&=&
-
\alpha {\bf Pat}
\underbrace{
-(\beta-\gamma)
c_{0101}
}_{\rm noise}=
-
\alpha\underbrace{b_1b_2}_{\rm Pat}
\underbrace{
-(\beta-\gamma)
b_2b_4
}_{\rm noise}={\bf Pat'}.
\ee
It remains to employ clean-up memory. But this is easy since the noise is perpendicular to
${\bf Pat}$. We only have to project on the set spanned by the fillers (i.e. the blades involving neither $b_3$ nor $b_4$), and within this set check which element is closest to the cleaned up
${\bf Pat'}$.

\section{Conclusions}

In order to switch from a binary string $x$, occurring in BSC, to HRR, one employs the exponential map $x\mapsto (-1)^{P_x}$ where $x$ plays a dual role. At the left-hand side $x$ is just a sequence of bits distributed over a $n$-tuple. At the right-hand side the bits are distributed over the diagonal of a diagonal $n\times n$ matrix $P_x$. Binary strings equipped with componentwise addition mod 2 (i.e. $\oplus$, XOR) form a group. The exponential map is a representation of this group in the space of diagonal matrices: $x\oplus y\mapsto (-1)^{P_x\oplus P_y}=(-1)^{P_x}(-1)^{P_y}$.

This group possesses also a projective representation in the space of blades of a GA: $x\oplus y\mapsto c_{x\oplus y}=\pm c_{x}c_{y}$. Blades have a straightforward geometric interpretation. As opposed to tensor products, that increase dimensions, the dimensions of $c_x$, $c_y$ and $c_{x\oplus y}$ are the same. Therefore, $c_{x\oplus y}$ can be used to bind variables. A superposition of such bound variables, a multivector, is a reduced representation analogous to HRR.

GA may also be interpreted as a way of encoding mutual geometric relations between multidimensional geometric objects. For example, a pair containing an oriented plane element and a vector from this plane is mapped in GA into a vector which shows how to move the first vector in order to produce the oriented plane segment in question. The algebraic operation thus reveals a geometric property of the plane segment, and resembles the process of {\it understanding\/} geometry.

We find the latter observation at least intriguing. It suggests that association of GA with cognitive science is not just a mathematical curiosity, but may be deeply rooted in the ways we think.

\acknowledgments
This work was supported by the Flemish Fund for Scientific Research (FWO), project G.0452.04.

\end{document}